%% file: arxiv_main.tex
\documentclass[10pt,twocolumn,letterpaper]{article}

\usepackage{cvpr}
\usepackage{times}
\usepackage{epsfig}
\usepackage{graphicx}
\usepackage{amsmath}
\usepackage{amssymb}

\usepackage{url}            %
\usepackage{booktabs}       %
\usepackage{amsfonts}       %
\usepackage{nicefrac}       %
\usepackage{color}
\usepackage{multirow}

\usepackage{placeins}
\usepackage[normalem]{ulem}
\usepackage[pagebackref=true,breaklinks=true,letterpaper=true,colorlinks,bookmarks=false]{hyperref}

\newcommand{\modelName}[1]{NOC}

\cvprfinalcopy %

\def\cvprPaperID{2454} %

\ifcvprfinal\fi
\begin{document}

\title{Captioning Images with Diverse Objects}

\author{
\vspace{-0.1cm} Subhashini Venugopalan$^\dagger$ \\
\and \vspace{-0.1cm} Lisa Anne Hendricks$^\ast$ \\
\and \vspace{-0.1cm} Marcus Rohrbach$^\ast$ \\
\and \hspace{-0.4cm}Raymond Mooney$^\dagger$ \\
\hspace{-0.8cm} $^\dagger$UT Austin\\
\hspace{-0.4cm}{\tt\footnotesize\{vsub,mooney\}@cs.utexas.edu}
\and \hspace{-1.2cm} Trevor Darrell$^\ast$\\
\hspace{-1.4cm}$^\ast$UC Berkeley\\ \vspace{-0.1cm}
\hspace{-0.5cm}{\tt\footnotesize \{lisa\_anne, rohrbach, trevor\}}\\
\hspace{-0.5cm}\tt\footnotesize{ @eecs.berkeley.edu}
\and \hspace{-0.8cm} Kate Saenko$^\ddagger$\\
\hspace{-0.8cm} $^\ddagger $Boston Univ.\\
\hspace{-0.8cm}{\tt\footnotesize saenko@bu.edu}
}

\maketitle
\begin{abstract}
Recent captioning models are limited in their ability to scale and describe concepts unseen in 
paired image-text corpora. We propose the Novel Object Captioner (\modelName{}), a deep visual semantic captioning model that can describe a large number of object categories not present in existing image-caption datasets. Our model takes advantage of external sources -- labeled images from object recognition datasets, and semantic knowledge extracted from unannotated text. 
We propose minimizing a joint objective which can learn from these diverse data sources and leverage distributional semantic embeddings,  enabling the model to generalize and describe novel objects outside of image-caption datasets. 
We demonstrate that our model exploits semantic information to generate captions for hundreds of object categories in the ImageNet object recognition dataset that are not observed in MSCOCO image-caption training data, as well as many categories that are observed very rarely. Both automatic evaluations and human judgements show that our model considerably outperforms prior work in being able to describe many more categories of objects.
\end{abstract}

\input{1intro}
\input{2related}

\input{4approach}
\input{5datasets}

\input{6experiments}

\input{7conclusion}
\vspace{-0.4cm}
\section*{Acknowledgements}
\vspace{-0.2cm}
We thank anonymous reviewers and Saurabh Gupta for helpful suggestions. Venugopalan is supported by a UT scholarship, and Hendricks was supported by a Huawei fellowship. Darrell was supported in part by DARPA; AFRL; DoD MURI award N000141110688; NSF awards  IIS-1212798, IIS-1427425, and IIS-1536003, and the Berkeley Artificial Intelligence Research Lab. Mooney and Saenko are supported in part by DARPA under AFRL grant FA8750-13-2-0026 and a Google Grant.
\input{8supplement}
\clearpage
\small
\bibliographystyle{ieee}
\bibliography{biblioShort,rohrbach,related,refs,nips_refs}

\end{document}

%% file: 1intro.tex
\vspace{-0.4cm}
\section{Introduction}
Modern visual classifiers~\cite{he16cvpr,vgg16arxiv} can recognize thousands of object categories, some of which are basic or entry-level (e.g. television), and others that are fine-grained and task specific (e.g. dial-phone, cell-phone).
However, recent state-of-the-art visual captioning systems~\cite{donahue15cvpr,fang15cvpr,karpathy15cvpr,kiros15tacl,mao15iclr,vinyals15cvpr} that learn directly from images and descriptions, rely solely on paired image-caption data for supervision and fail in their ability to generalize and describe this vast set of recognizable objects in context. 
While such systems could be scaled by building larger image/video description datasets, obtaining such captioned data would be expensive and laborious.
Furthermore, visual description is challenging because models have to not only correctly identify visual concepts contained in an image, but must also 
compose these concepts into a coherent sentence.

\begin{figure}[t]
\begin{center}
\includegraphics[width=\linewidth]{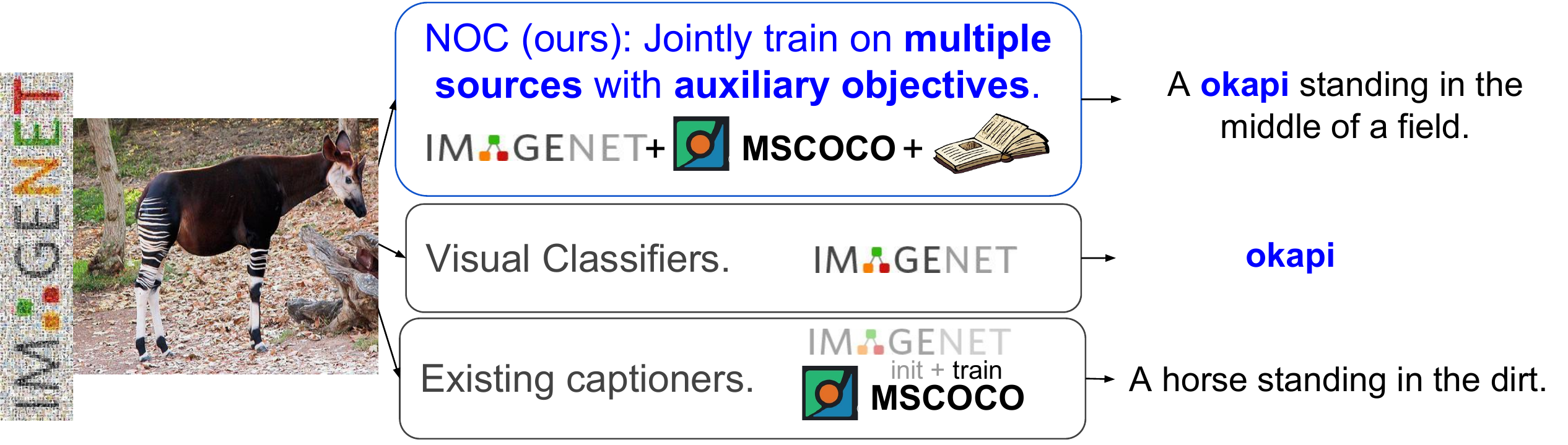}
\end{center}
\vspace{-0.4cm}
 \caption{\small We propose a model that learns simultaneously from multiple data sources with auxiliary objectives to describe a variety of objects unseen in paired image-caption data.
}
\label{fig:noc-teaser}
\end{figure}

Recent work \cite{hendricks16cvpr} shows that, to incorporate the vast knowledge of current visual recognition networks without explicit paired caption training data, caption models can learn from external sources and learn to compose sentences about visual concepts which are infrequent or non-existent in image-description corpora. 
However, the pioneering DCC model from \cite{hendricks16cvpr} is unwieldy in the sense that the model requires explicit transfer (``copying'') of learned parameters from previously seen categories to novel categories. This not only prevents it from describing rare categories and limits the model's ability to cover a wider variety of objects but also makes it unable to be trained end-to-end.
We instead propose the Novel Object Captioner (\modelName{}), a network that can be trained end-to-end using a joint training strategy to integrate knowledge from external visual recognition datasets as well as semantic information from independent unannotated text corpora to generate captions for a diverse range of rare and novel objects (as in Fig.~\ref{fig:noc-teaser}). %

Specifically, we introduce auxiliary objectives which allow our network to learn a captioning model on image-caption pairs simultaneously with a deep language model and visual recognition system on unannotated text and labeled images.
Unlike previous work, the auxiliary objectives allow the \modelName{} model to learn relevant information from multiple data sources simultaneously in an end-to-end fashion. 
Furthermore, \modelName{} implicitly leverages pre-trained distributional word embeddings enabling it to describe unseen and rare object categories. The main contributions of our work are $1)$ an end-to-end model to describe objects not present in paired image-caption data, $2)$ auxiliary/joint training of the visual and language models on multiple data sources, and $3)$ incorporating pre-trained semantic embeddings for the task.
We demonstrate the effectiveness of our model by performing extensive experiments on objects held out from MSCOCO \cite{coco2014} as well as hundreds of objects from ImageNet \cite{imagenet2014} unseen in caption datasets. Our model substantially outperforms previous work \cite{hendricks16cvpr} on both automated as well as human evaluations.

%% file: 2related.tex
\section{Related Work}
\paragraph{Visual Description.} 
This area has seen many different approaches over the years \cite{daume11generation,kuznetsova14tacl,mitchell12eacl},
and more recently deep models have gained popularity for both their performance and potential for end-to-end training.
Deep visual description frameworks first encode an image into a fixed length feature vector and then generate a description by either conditioning text generation on image features \cite{donahue15cvpr,karpathy15cvpr,vinyals15cvpr} or embedding image features and previously generated words into a multimodal space \cite{kiros2014multimodal,kiros15tacl,mao15iclr} before predicting the next word.
Though most models represent images with an intermediate representation from a convolutional neural network (such as fc$_{7}$ activations from a CNN), 
other models represent images as a vector of confidences over a fixed number of visual concepts \cite{fang15cvpr,hendricks16cvpr}. In almost all cases, the parameters of the visual pipeline are initialized with weights trained on the ImageNet classification task.
For caption generation, recurrent networks (RNNs) are a popular choice to model language, but log bilinear models \cite{kiros2014multimodal} and maximum entropy language models \cite{fang15cvpr} have also been explored. Our model is similar to the CNN-RNN frameworks in \cite{hendricks16cvpr,mao15iclr} but neither of these models can be trained end-to-end to describe objects unseen in image-caption pairs.

\vspace{-0.4cm}
\paragraph{Novel object captioning.}
\cite{mao15iccv} proposed an approach that extends a model's capability to describe a small set of novel concepts (e.g. {\it{quidditch, samisen}}) from a few paired training examples while retaining its ability to describe previously learned concepts. On the other hand, \cite{hendricks16cvpr} introduce a model that can describe many objects already existing in English corpora and object recognition datasets (ImageNet) but not in the caption corpora (e.g. {\it{pheasant, otter}}). Our focus is on the latter case. \cite{hendricks16cvpr} integrate information from external text and visual sources, and explicitly transfer (`copy') parameters from objects seen in image-caption data to unseen ImageNet objects to caption these novel categories. While this works well for many ImageNet classes it still limits coverage across diverse categories
and cannot be trained end-to-end. Furthermore, their model cannot caption objects for which few paired training examples already exist. Our proposed framework integrates distributional semantic embeddings implicitly, obviating the need for any explicit transfer and making it end-to-end trainable. It also extends directly to caption ImageNet objects with few or no descriptions.

\vspace{-0.2cm}
\paragraph{Multi-modal and Zero-Shot Learning.} 
Another closely related line of research takes advantage of distributional semantics to learn a joint embedding space using visual and textual information for zero-shot labeling of novel object categories \cite{frome2013devise,norouzi2013conse}, as well as retrieval of images with text \cite{lazaridou2014wampimuk,socher14tacl}.
Visual description itself can be cast as a multimodal learning problem in which caption words $w_0, ..., w_{n-1}$ and an image are projected into a joint embedding space before the next word in a caption, $w_n$, is generated \cite{kiros15tacl,mao15iclr}.
Although our approach uses distributional word embeddings, our model differs in the sense that it can be trained with unpaired text and visual data but still combine the semantic information at a later stage during caption generation.
This is similar in spirit to works in natural language processing that use monolingual data to improve machine translation \cite{gulcehreArxiv15}. %

%% file: 4approach.tex
\vspace{-0.2cm}
\section{Novel Object Captioner (NOC)}
\vspace{-0.2cm}

\begin{figure}[t]
\begin{center}
\includegraphics[width=\linewidth]{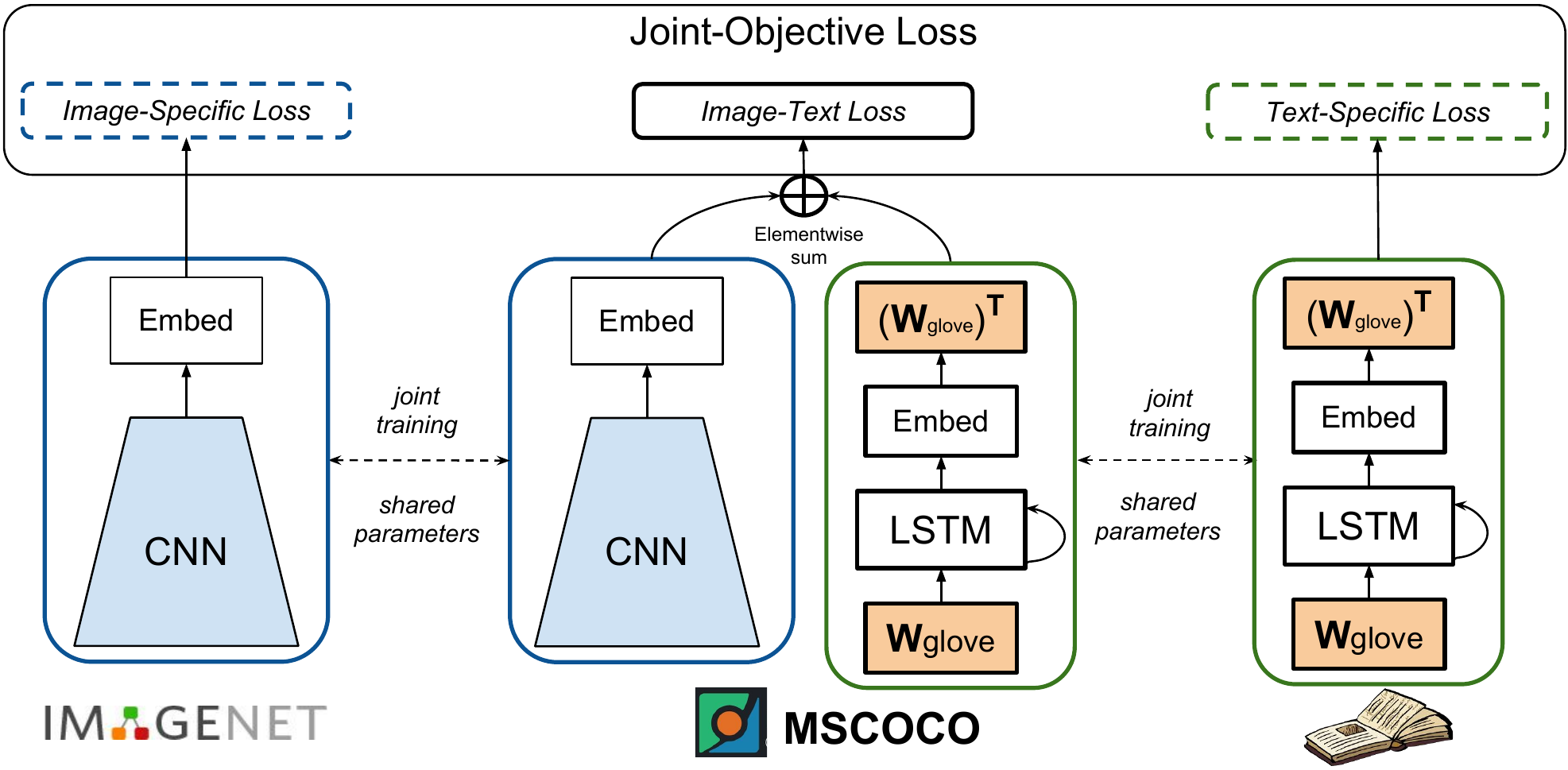}
\end{center}
\vspace{-0.3cm}
 \caption{\small
Our \modelName{} image caption network. During training, the visual recognition network (left), the LSTM-based language model (right), and the caption model (center) are trained simultaneously on different sources with different objectives but with shared parameters, thus enabling novel object captioning. %
}
\label{fig:noc}
\end{figure}

Our \modelName{} model is illustrated in Fig.~\ref{fig:noc}.
It consists of a language model that leverages distributional semantic embeddings trained on unannotated text and integrates it with a visual recognition model. 
We introduce auxiliary loss functions (objectives) and jointly train different components on multiple data sources, to create a visual description model which simultaneously learns an independent object recognition model, as well as a language model.

We start by first training a LSTM-based language model (LM) \cite{sundermeyer12inter} for sentence generation. Our LM incorporates dense representations for words from distributional embeddings (GloVe, \cite{Glove}) pre-trained on external text corpora. %
Simultaneously, we also train a state-of-the-art visual recognition network to provide confidences over words in the vocabulary given an image. This decomposes our model into discrete textual and visual pipelines which can be trained exclusively using unpaired text and unpaired image data (networks on left and right of Fig.~\ref{fig:noc}).
To generate descriptions conditioned on image content, we combine the predictions of our language and visual recognition networks by summing (element-wise) textual and visual confidences over the vocabulary of words.
During training, we introduce auxiliary image-specific ($\mathcal{L_{IM}}$), and text-specific ($\mathcal{L_{LM}}$) objectives along with the paired image-caption ($\mathcal{L_{CM}}$) loss. These loss functions, when trained jointly, influence our model to not only produce reasonable image descriptions, but also predict visual concepts as well as generate cohesive text (language modeling).
We first discuss the auxiliary objectives and the joint training,
and then discuss how we leverage embeddings trained with external text to compose descriptions about novel objects.

\vspace{-0.1cm}
\subsection{Auxiliary Training Objectives}
\vspace{-0.1cm}
Our motivation for introducing auxiliary objectives is to learn how to describe images without losing the ability to recognize more objects.
Typically, image-captioning models incorporate a visual classifier pre-trained on a source domain (e.g. ImageNet dataset) and then tune it to the target domain (the image-caption dataset). However, important information from the source dataset can be suppressed if similar information is not present when fine-tuning, leading the network to forget (over-write weights) for objects not present in the target domain. This is problematic in our scenario in which the model relies on the source datasets to learn a large variety of visual concepts not present in the target dataset. However, with pre-training as well as the complementary auxiliary objectives the model maintains its ability to recognize a wider variety of objects and is encouraged to describe objects which are not present in the target dataset at test time. 
For the ease of exposition, we abstract away the details of the language and the visual models and first describe the joint training objectives of the complete model, i.e. the text-specific loss, the image-specific loss, and the image-caption loss. We will then describe the language and the visual models.

\vspace{-0.3cm}
\newcommand{\softmax}[2]{S_{#1}(#2)}
\subsubsection{Image-specific Loss}
\vspace{-0.1cm}
Our visual recognition model (Fig.~\ref{fig:noc}, left) is a neural network parametrized by $\theta_I$ and is trained on object recognition datasets. Unlike typical visual recognition models that are trained with a single label on a classification task, for the task of image captioning an image model that has high confidence over multiple visual concepts occurring in an image simultaneously would be preferable. Hence, we choose to train our model using multiple labels (more in Sec.~\ref{subsec:mscoco_heldout1}) with a multi-label loss. If $l$ denotes a label and $z_l$ denotes the binary ground-truth value for the label, then the objective for the visual model is given by the cross-entropy loss ($\mathcal{L}_{IM}$):
\vspace{-0.2cm}
\begin{align}
\label{eqn:imL}
\begin{split}
\vspace{-0.2cm}
\mathcal{L_{IM}}(I; \theta_{I}) &= - \sum\limits_{l} \Bigl[ z_l \  log(\softmax{l}{f_{IM}(I; \theta_{I})}) \\
  &+ (1-z_l)\ log(1- \softmax{l}{f_{IM}(I; \theta_{I})}) \Bigr]
  \end{split}
\end{align}
where $\softmax{i}{x}$ is the output of a $\mathrm{softmax}$ function over index $i$ and input $x$, and $f_{IM}$, is the activation of the final layer of the visual recognition network.

\vspace{-0.3cm}
\subsubsection{Text-specific Loss}
\vspace{-0.1cm}
Our language model (Fig.~\ref{fig:noc}, right) is based on LSTM Recurrent Neural Networks. We denote the parameters of this network by $\theta_L$, and the activation of the final layer of this network by $f_{LM}$. The language model is trained to predict the next word $w_t$ in a given sequence of words $w_0, ..., w_{t-1}$. This is optimized using the $\mathrm{softmax}$ loss $\mathcal{L_{LM}}$ which is equivalent to the maximum-likelihood:
\vspace{-0.1cm}
\begin{align}
\begin{split}
 &\mathcal{L_{LM}}(w_0,...,w_{t-1}; \theta_{L}) = \\
& - \sum\limits_{t} log(\softmax{w_t}{f_{LM}(w_0,...,w_{t-1}; \theta_{L})})  
\end{split}
\end{align}

\vspace{-0.6cm}
\subsubsection{Image-caption Loss}
\vspace{-0.1cm}
The goal of the image captioning model (Fig.~\ref{fig:noc}, center) is to generate a sentence conditioned on an image ($I$).  \modelName{} predicts the next word in a sequence, $w_t$, conditioned on previously generated words ($w_0, ..., w_{t-1}$) and an image ($I$), by summing activations from the deep language model, which operates over previous words, and the deep image model, which operates over an image. We denote these final (summed) activations by $f_{CM}$. Then, the probability of predicting the next word is given by, $P(w_t| w_0, ..., w_{t-1}, I)$
\begin{align}
\vspace{-0.4cm}
\begin{split}
=& \softmax{w_t}{f_{CM}(w_0, ..., w_{t-1}, I; \theta)} \\
=& \softmax{w_t}{f_{LM}(w_0, ..., w_{t-1}; \theta_L) + f_{IM}(I; \theta_I)}
\end{split}
\end{align}
Given pairs of images and descriptions, the caption model optimizes the parameters of the underlying language model ($\theta_L$) and image model ($\theta_I$) by minimizing the caption model loss $\mathcal{L_{CM}}$ : $\mathcal{L_{CM}}(w_0,.,w_{t-1},I; \theta_{L}, \theta_{I})$
\vspace{-0.2cm}
\begin{align}
\label{eqn:cm}
= -\sum\limits_{t} log(\softmax{w_t}{{f_{CM}(w_0,.,w_{t-1},I; \theta_{L}, \theta_{I})}})  
\end{align}

\vspace{-0.5cm}
\subsubsection{Joint Training with Auxiliary Losses}
\vspace{-0.1cm}
While many previous approaches have been successful on image captioning by pre-training the image and language models and tuning the caption model alone (Eqn.~\ref{eqn:cm}), this is insufficent to generate descriptions for objects outside of the image-caption dataset since the model tends to ``forget'' (over-write weights) for objects only seen in external data sources. To remedy this, we propose to train the image model, language model, and caption model simultaneously on different data sources. The \modelName{} model's final objective simultaneously minimizes the three individual complementary objectives:
\vspace{-0.2cm}
\begin{align}
\vspace{-0.2cm}
\label{eqn:jointL}
\mathcal{L} = \mathcal{L_{CM}} + \mathcal{L_{IM}} + \mathcal{L_{LM}}
\end{align}
By sharing the weights of the caption model's network with the image network and the language network (as depicted in Fig.~\ref{fig:noc} (a)), the model can be trained simultaneously on independent image-only data, unannotated text data, as well as paired image-caption data. Consequently, co-optimizing different objectives aids the model in recognizing categories outside of the paired image-sentence data.

\subsection{Language Model with Semantic Embeddings}
\vspace{-0.1cm}
\label{sec:approach:LM}
Our language model consists of the following components: a continuous lower dimensional embedding space for words ($W_{glove}$), a single recurrent (LSTM) hidden layer, and two linear transformation layers where the second layer ($W_{glove}^T$) maps the vectors to the size of the vocabulary.
Finally a softmax activation function is used on the output layer to produce a normalized probability distribution. The cross-entropy loss which is equivalent to the maximum-likelihood is used as the training objective. %
In addition to our joint objective (Eqn.\ref{eqn:jointL}), we also employ semantic embeddings in our language model to help generate sentences when describing novel objects.
Specifically, the initial input embedding space ($W_{glove}$) is used to represent the input (one-hot) words into semantically meaningful dense fixed-length vectors. While the final transformation layer ($W^T_{glove}$) 
reverses the mapping \cite{mao15iclr,venugopalan16emnlp} of a dense vector back to the full vocabulary with the help of a $\mathrm{softmax}$ activation function. 
These distributional embeddings \cite{mikolov13nips,Glove} share the property that words that are semantically similar have similar vector representations. The intuitive reason for using these embeddings in the input and output transformation layers is to help the language model treat words unseen in the image-text corpus to (semantically) similar words that have previously been seen so as to encourage compositional sentence generation i.e. encourage it to use new/rare word in a sentence description based on the visual confidence.

\vspace{-0.1cm}
\subsection{Visual Classifier}
\vspace{-0.1cm}
\label{sec:vismodel}
The other main component of our model is the visual classifier. Identical to previous work \cite{hendricks16cvpr}, 
we employ the VGG-16~\cite{vgg16arxiv} convolutional network as the visual recognition network. %
We modify the final layers of the network to incorporate the multi-label loss (Eqn.~\ref{eqn:imL}) to predict visual confidence over multiple labels in the full vocabulary. The rest of the classification network remains unchanged.

Finally, we take an elementwise-sum of the visual and language outputs, one can think of this as the language model producing a smooth probability distribution over words (based on GloVe parameter sharing) and then the image signal ``selecting'' among these based on the visual evidence when summed with the language model beliefs.

%% file: 5datasets.tex
\vspace{-0.2cm}
\section{Datasets}
\vspace{-0.1cm}
In this section we describe the image description dataset as well as the external text and image datasets used in our experiments. 

\vspace{-0.1cm}
\subsection{External Text Corpus (WebCorpus)}
\vspace{-0.2cm}
We extract sentences from Gigaword, the British National Corpus (BNC), UkWaC, and Wikipedia. Stanford CoreNLP 3.4.2 \cite{corenlp} was used to extract tokenizations. This dataset was used to train the LSTM language model. For the dense word representation in the network, we use GloVe \cite{Glove} pre-trained on 6B tokens of external corpora including Gigaword and Wikipedia. To create our LM vocabulary we identified the 80,000 most frequent tokens from the combined external corpora. We refine this vocabulary further to a set of 72,700 words that also had GloVe embeddings.

\vspace{-0.2cm}
\subsection{Image Caption data} 
\label{sec:img_data}
\vspace{-0.2cm}
To empirically evaluate the ability of \modelName{} to describe new objects we use the training and test set from \cite{hendricks16cvpr}. 
This dataset is created from MSCOCO \cite{coco2014} by clustering the main 80 object categories using cosine distance on word2vec (of the object label) and selecting one object from each cluster to hold out from training. The training set holds out images and sentences of 8 objects (bottle, bus, couch, microwave, pizza, racket, suitcase, zebra), which constitute about 10$\%$ of the training image and caption pairs in the MSCOCO dataset.
Our model is evaluated on how well it 
can generate descriptions about images containing the eight held-out objects.

\begin{table*}[!htbp]
\small
\begin{center}
\begin{tabular}{lccccccccc| c}
\toprule
 Model &  bottle & bus & couch & microwave & pizza & racket & suitcase & zebra & Avg. F1 & {Avg. \footnotesize{METEOR}} \\
\cmidrule(lr){1-9}\cmidrule(lr){10-10}\cmidrule(lr){11-11}
DCC & 4.63 & 29.79 & \textbf{45.87} & \textbf{28.09} & 64.59 & 52.24 & 13.16 & 79.88 & 39.78 & 21.00 \\
\modelName{} (ours) & \textbf{17.78} & \textbf{68.79} & 25.55 & 24.72 & \textbf{69.33} & \textbf{55.31} & \textbf{39.86 }& \textbf{89.02} & {\bf{48.79}} & \textbf{21.32}\\
\bottomrule
\end{tabular}
\end{center}
\caption{\small
MSCOCO Captioning: F1 scores (in \%) of \modelName{} (our model) and DCC \protect\cite{hendricks16cvpr} on held-out objects not seen jointly during image-caption training, along with the average F1 and METEOR scores of the generated captions across images containing these objects.%
}
\label{tab:results:coco-8objs}
\vspace{-2.0mm}
\end{table*}

\vspace{-0.2cm}
\subsection{Image data}
\vspace{-0.2cm}
We also evaluate sentences generated by \modelName{} on approximately 700 different ImageNet \cite{imagenet2014} objects which are not present in the MSCOCO dataset. We choose this set by identifying objects that are present in both ImageNet and our language corpus (vocabulary), but not present in MSCOCO.
Chosen words span a variety of categories including fine-grained categories (e.g., ``bloodhound'' and ``chrysanthemum''), adjectives (e.g., ``chiffon'', ``woollen''), and entry level words (e.g., ``toad''). 
Further, to study how well our model can describe rare objects, we pick a separate set of 52 objects which are in ImageNet but mentioned infrequently in MSCOCO (52 mentions on average, with median 27 mentions across all 400k training sentences).

%% file: 6experiments.tex
\vspace{-0.2cm}
\section{Experiments on MSCOCO}
\vspace{-0.1cm}

We perform the following experiments to compare \modelName{}'s performance with previous work \cite{hendricks16cvpr}:
1.~We evaluate the model's ability to caption objects that are held out from MSCOCO during training  (Sec.~\ref{subsec:mscoco_heldout1}).
2.~To study the effect of the data source on training, we report performance of \modelName{} when the image and language networks are trained on in-domain and out-of-domain sources (Sec.~\ref{subsec:exp_mscoco_datasource}).
In addition to these, to understand our model better:
3.~We perform ablations to study how much each component of our model (such as word embeddings, auxiliary objective, etc.) contributes to the performance (Sec.~\ref{subsec:exp_mscoco_ablations}).
4.~We also study if the model's performance remains consistent when holding out a different subset of objects from MSCOCO (Sec.~\ref{subsec:exp_mscoco_heldoutnew}).

\vspace{-0.1cm}
\subsection{Empirical Evaluation on MSCOCO} \label{subsec:mscoco_heldout1}
\vspace{-0.2cm}
We empirically evaluate the ability of our proposed model to describe novel objects by following the experimental setup of \cite{hendricks16cvpr}.
We optimize each loss in our model with the following datasets: the caption model, which jointly learns the parameters $\theta_L$ and $\theta_I$, is trained only on the subset of MSCOCO without the 8 objects (see section~\ref{sec:img_data}), the image model, which updates parameters $\theta_I$, is optimized using labeled images, and the language model which updates parameters $\theta_L$, is trained using the corresponding descriptions. When training the visual network on images from COCO, we obtain multiple labels for each image by considering all words in the associated captions as labels after removing stop words. We first present evaluations for the in-domain setting in which the image classifier is trained with all COCO training images and the language model is trained with all sentences. %
We use the METEOR metric \cite{lavie2014meteor} to evaluate description quality. However, METEOR only captures fluency and does not account for the mention (or lack) of specific words. Hence, we also use F1 to ascertain that the model mentions the object name in the description of the images containing the object. Thus, the metrics measure if the model can both identify the object and use it fluently in a sentence.

\begin{figure}[]
\vspace{-0.3cm}
\begin{center}
\includegraphics[scale=0.6]{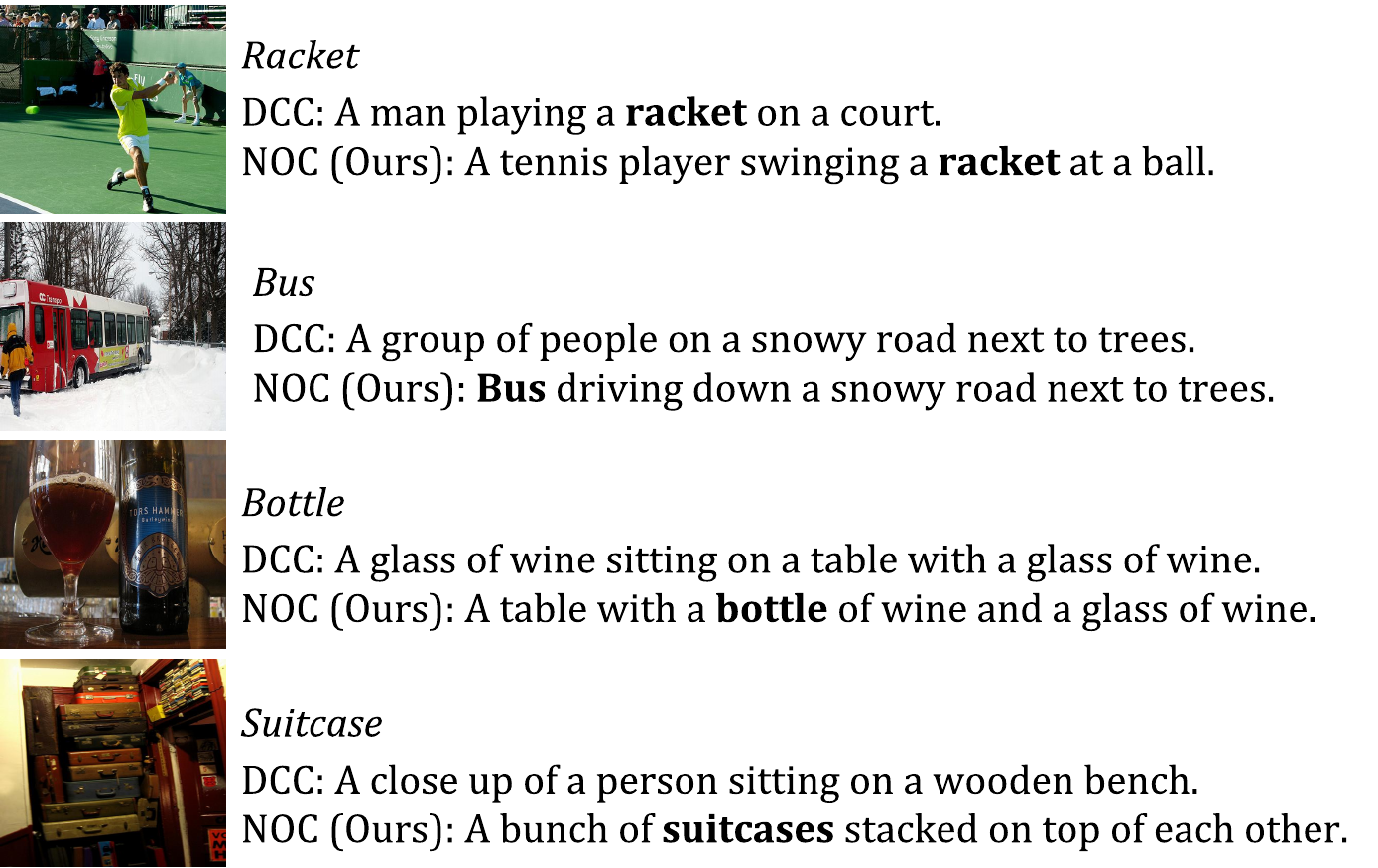}
\end{center}
\vspace{-0.2cm}
    \caption{COCO Captioning: Examples comparing captions by NOC (ours) and DCC \cite{hendricks16cvpr} on held out objects from MSCOCO.}
    \label{fig:coco_dcc_noc}
\end{figure}

\textbf{COCO heldout objects.}
Table~\ref{tab:results:coco-8objs} compares the F1  score achieved by \modelName{} to the previous best method, DCC~\cite{hendricks16cvpr} on the 8 held-out COCO objects.
\modelName{} outperforms DCC (by 10\% F1 on average) on all objects except ``couch'' and ``microwave''.
The higher F1 and METEOR demonstrate that \modelName{} is able to correctly recognize many more instances of the unseen objects and also integrate the words into fluent descriptions. 

\begin{table}
\small
\begin{center}
\begin{tabular}{lccccc}
\toprule
& Image & Text & Model & {\footnotesize METEOR} & F1 \\
\cmidrule(lr){1-6}
\multirow{2}{*}{1} & \multicolumn{2}{c}{Baseline} & LRCN & 19.33& 0 \\
 & \multicolumn{2}{c}{({\small no transfer})} & DCC & 19.90 & 0 \\
\cmidrule(lr){1-6}
\multirow{2}{*}{2} & {\footnotesize{Image}} & {\footnotesize Web} &  DCC  & 20.66 & 34.94 \\
 &{\footnotesize{Net}} & {\footnotesize Corpus} & \modelName{} & 17.56 & 36.50 \\
 \cmidrule(lr){1-6}
\multirow{2}{*}{3} & \multirow{2}{*}{\footnotesize COCO} & {\footnotesize Web} & \multirow{2}{*}{\modelName{}} & \multirow{2}{*}{19.18} & \multirow{2}{*}{41.74} \\
 & & {\footnotesize Corpus} & & \\
\cmidrule(lr){1-6}  
\multirow{2}{*}{4} & \multirow{2}{*}{\footnotesize COCO} & \multirow{2}{*}{\footnotesize COCO} & DCC  & {21.00} & 39.78 \\
& & & \modelName{} & \textbf{21.32} & \textbf{48.79} \\
\bottomrule
\end{tabular}
\end{center}
\caption{\small Comparison with different training data sources on 8 held-out COCO objects. Having in-domain data helps both the DCC \protect\cite{hendricks16cvpr} and our \modelName{} model caption novel objects.}
\label{tab:noc-comp}
\end{table}

\vspace{-0.1cm}
\subsection{Training data source}\label{subsec:exp_mscoco_datasource}
\vspace{-0.1cm}
To study the effect of different data sources,
we also evaluate our model in an out-of-domain setting where classifiers for held out objects are trained with images from ImageNet and the language model is trained on text mined from external corpora.
Table~\ref{tab:noc-comp} reports average scores across the eight held-out objects.
We compare our \modelName{} model to results from \cite{hendricks16cvpr} (DCC), as well as a competitive image captioning model - LRCN~\cite{donahue15cvpr} trained on the same split. %
In the out-of-domain setting (line 2), for the chosen set of 8 objects, \modelName{} performs slightly better on F1 and a bit lower on METEOR compared to DCC. 
However, as previously mentioned, DCC needs to explicitly identify a set of ``seen'' object classes to transfer weights to the novel classes whereas \modelName{} can be used for inference directly.
DCC's transfer mechanism %
also leads to peculiar descriptions. E.g., \textit{Racket} in Fig.~\ref{fig:coco_dcc_noc}.

With COCO image training (line 3), F1 scores of \modelName{} improves considerably even with the Web Corpus LM training. Finally in the in-domain setting (line 4) \modelName{} outperforms DCC on F1 by around 10 points while also improving METEOR slightly.
This suggests that \modelName{} is able to associate the objects with captions better with in-domain training, and the auxiliary objectives and embedding help the model to generalize and describe novel objects.

\begin{table*}[t]
\begin{center}
\begin{tabular}{|c| cccc | c | c |}
  \hline
 \multirow{2}{*}{Contributing factor} & \multirow{2}{*}{Glove} & \multirow{2}{*}{LM pretrain}  & Tuned Visual & Auxiliary  &  \multirow{2}{*}{\footnotesize{METEOR}} & \multirow{2}{*}{F1} \\ 
      &  &   & Classifier & Objective & & \\\hline
Tuned Vision & - & - & \checkmark &\checkmark & 15.78 & 14.41 \\
LM \& Embedding &  \checkmark & \checkmark & \checkmark & - & 19.80 & 25.38 \\
LM \& Pre-trained Vision & \checkmark & \checkmark & Fixed & -  & 18.91 & 39.70 \\
Auxiliary Objective & \checkmark & - & \checkmark & \checkmark & 19.69 & 47.02 \\
 All &  \checkmark & \checkmark & \checkmark&\checkmark & \textbf{21.32} & \textbf{48.79} \\ \hline
\end{tabular}
\end{center}
\caption{Ablations comparing the contributions of the Glove embedding, LM pre-training, and auxiliary objectives, of the NOC model.  Our auxiliary objective along with Glove have the largest impact in captioning novel objects.}
\label{tab:ablations}
\end{table*}

\begin{table*}[htb]
\small
\begin{center}
\begin{tabular}{lccccccccc| c}
\toprule
 Model &  bed & book & carrot & elephant & spoon & toilet & truck & umbrella & Avg. F1 & {Avg. \footnotesize{METEOR}} \\
\cmidrule(lr){1-9}\cmidrule(lr){10-10}\cmidrule(lr){11-11}
\modelName{} (ours) & {53.31} & {18.58} & 20.69 & 85.35 & {02.70} & {73.61} & {57.90 }& {54.23} & {{45.80}} & {20.04}\\
\bottomrule
\end{tabular}
\end{center}
\caption{\small
MSCOCO Captioning: F1 scores (in \%) of \modelName{} (our model) on a different subset of the held-out objects not seen jointly during image-caption training, along with the average F1 and METEOR scores of the generated captions across images containing these objects.  NOC is consistently able to caption different subsets of unseen object categories in MSCOCO.  %
}
\label{tab:results:coco-new8objs}
\vspace{-2.0mm}
\end{table*}

\subsection{Ablations} \label{subsec:exp_mscoco_ablations}
Table~\ref{tab:ablations} compares how different aspects of training impact the overall performance. 
{\textit{Tuned Vision contribution}} The model that does not incorporate Glove or LM pre-training has poor performance (METEOR 15.78, F1 14.41); this ablation shows the contribution of the vision model alone in recognizing and describing the held out objects. %
{\textit{LM \& Glove contribution:}} The model trained without the auxiliary objective, performs better with F1 of 25.38 and METEOR of 19.80; this improvement comes largely from the GloVe embeddings which help in captioning novel object classes.
 {\textit{LM \& Pre-trained Vision:}} It's interesting to note that when we fix classifier's weights (pre-trained on all objects), before tuning the LM on the image-caption COCO subset, the F1 increases substantially to 39.70 suggesting that the visual model recognizes many objects but can ``forget'' objects learned by the classifier when fine-tuned on the image-caption data (without the 8 objects).
 {\textit{Auxiliary Objective:}} Incorporating the auxiliary objectives, F1 improves remarkably to 47.02. We note here that by virtue of including auxiliary objectives the visual network is tuned on all images thus retaining it's ability to classify/recognize a wide range of objects.
Finally, incorporating all aspects gives \modelName{} the best performance (F1 48.79, METEOR 21.32), significantly outperforming DCC.

\subsection{Validating on a different subset of COCO} \label{subsec:exp_mscoco_heldoutnew}
To show that our model is consistent across objects, we create a different training/test split by holding out a different set of eight objects from COCO. The objects we hold out are: bed, book, carrot, elephant, spoon, toilet, truck and umbrella. Images and sentences from these eight objects again constitute about 10\% of the MSCOCO training dataset. Table \ref{tab:results:coco-new8objs} presents the performance of the model on this subset. We observe that the F1 and METEOR scores, although a bit lower, are consistent with numbers observed in Table \ref{tab:results:coco-8objs} confirming that our model is able to generalize to different subsets of objects.

\section{Experiments: Scaling to ImageNet}
To demonstrate the scalability of \modelName{}, we describe objects in ImageNet for which no paired image-sentence data exists. Our experiments are performed on two subsets of ImageNet, (i) Novel Objects: A set of 638 objects which are present in ImageNet as well as the model's vocabulary but are not mentioned in MSCOCO. (ii) Rare Objects: A set of 52 objects which are in ImageNet as well as the MSCOCO vocabulary but are mentioned infrequently in the MSCOCO captions (median of 27 mentions). For quantitative evaluation, (i) we measure the percentage of objects for which the model is able to describe at least one image of the object (using the object label), (ii) we also report accuracy and F1 scores to compare across the entire set of images and objects the model is able to describe. Furthermore, we obtain human evaluations comparing our model with previous work on whether the model is able to incorporate the object label meaningfully in the description together with how well it describes the image.

\subsection{Describing Novel Objects}
Table~\ref{tab:imgnet-stats} compares models on 638 novel object categories (identical to \cite{hendricks16cvpr}) using the following metrics: (i) Describing novel objects (\%) refers to the percentage of the selected ImageNet objects mentioned in descriptions, i.e.\ for each novel word (e.g., ``otter'') the model should incorporate the word (``otter'') into at least one description about an ImageNet image of the object (otter).
While DCC is able to recognize and describe 56.85\% (363) of the selected ImageNet objects in descriptions, \modelName{} recognizes several more objects and is capable of describing 91.27\% (582 of 638) ImageNet objects.
(ii) Accuracy refers to the percentage of images from each category where the model is able to correctly identify and describe the category. We report the average accuracy across all categories.
DCC incorporates a new word correctly 11.08\% of the time, in comparison, \modelName{} improves this appreciably to 24.74\%.
(iii) F1 score is computed based on precision and recall of mentioning the object in the description. Again, \modelName{} outperforms with average F1 33.76\%  to DCC's 14.47\%.

Although \modelName{} and DCC~\cite{hendricks16cvpr} use the same CNN, \modelName{} is both able to describe more categories, and correctly integrate new words into descriptions more frequently. 
DCC~\cite{hendricks16cvpr} can fail either with respect to finding a suitable object that is both semantically and syntactically similar to the novel object, or with regard to their language model composing a sentence using the object name, in \modelName{} the former never occurs (i.e.\ we don't need to explicitly identify similar objects), reducing the overall sources of error. %
\begin{table}[t]
\vspace{-0.2cm}
\begin{center}
\small
\begin{tabular}{| l | c | c | c |}
  \hline
   Model & Desc. Novel (\%) &  Acc (\%) & F1 (\%) \\ \hline
  DCC & 56.85 &  11.08 & 14.47  \\
  \modelName{} & \textbf{91.27} & \textbf{24.74} & \textbf{33.76} \\
  \hline
\end{tabular}
\end{center}
\caption{ImageNet: Comparing our model against DCC \protect\cite{hendricks16cvpr} on \% of novel classes described, average accuracy of mentioning the class in the description, and mean F1 scores for object mentions.}
\label{tab:imgnet-stats}
\end{table}

\begin{figure}[]
\vspace{-0.3cm}
\begin{center}
\includegraphics[scale=0.5]{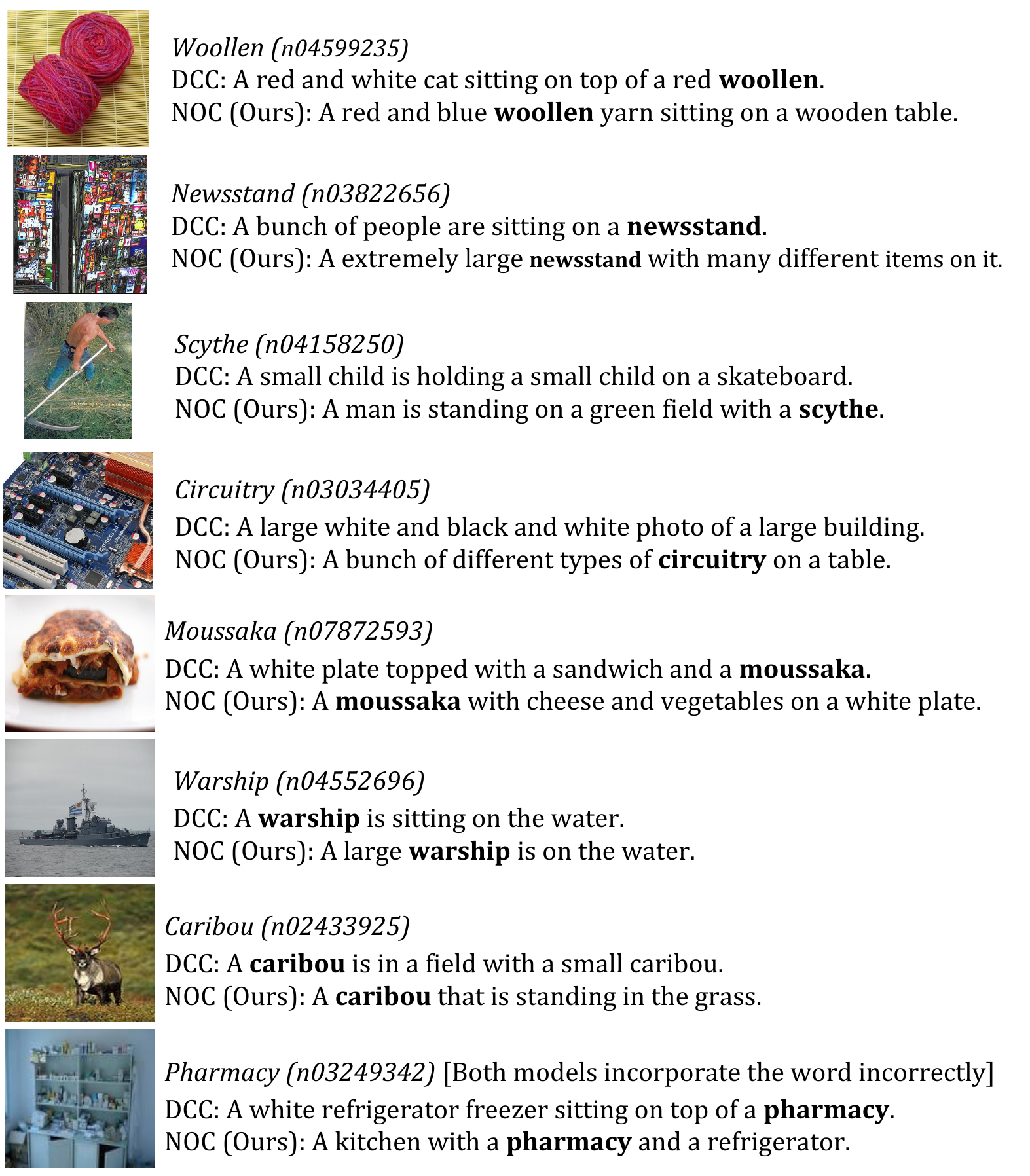}
\end{center}
    \caption{ImageNet Captioning: Examples comparing captions by NOC (ours) and DCC \cite{hendricks16cvpr} on objects from ImageNet.}
    \label{fig:imnet_dcc_noc}
\end{figure}

\begin{figure}[]
\vspace{-0.3cm}
\begin{center}
\includegraphics[scale=0.6]{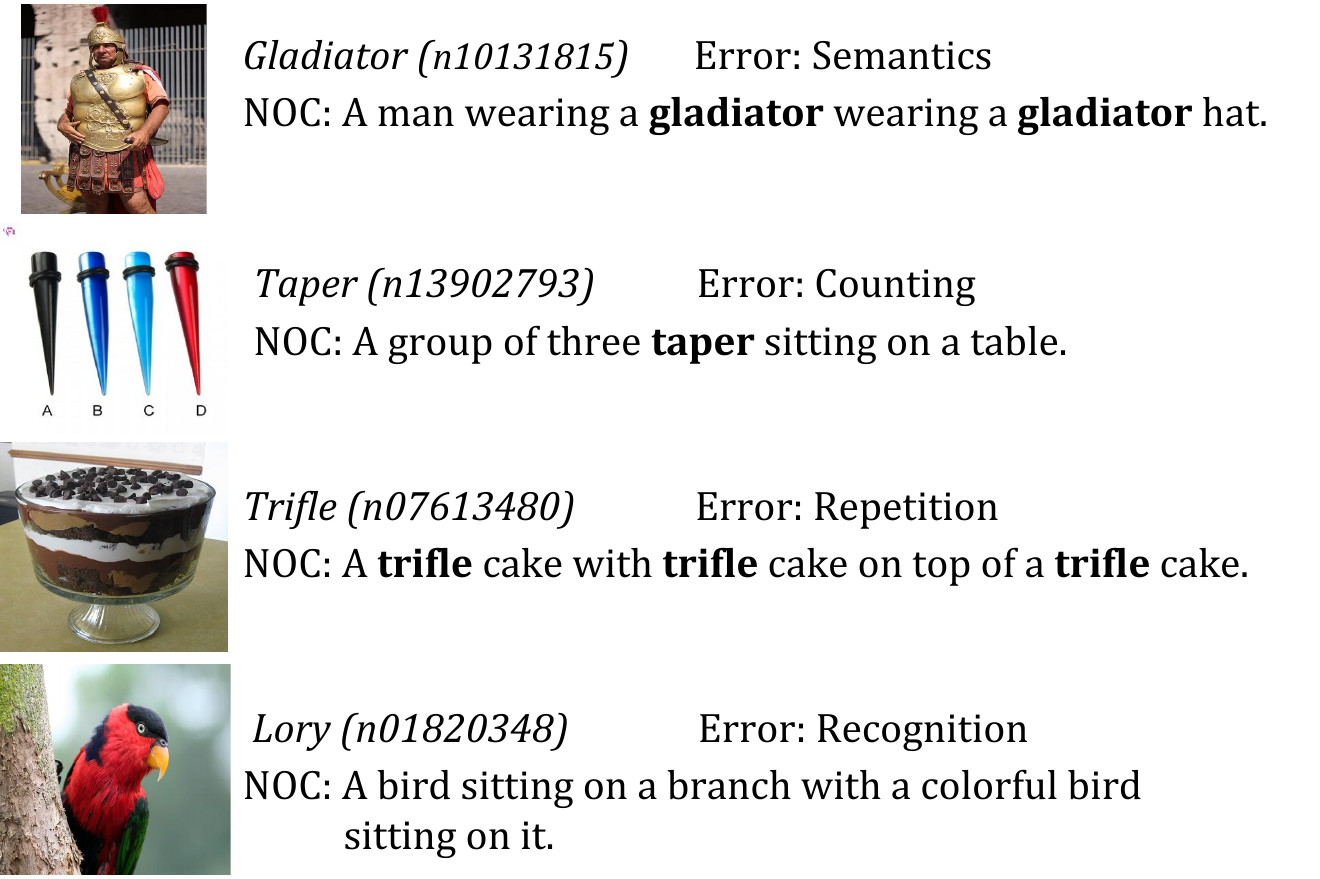}
\end{center}
    \caption{ImageNet Captioning: Common types of errors observed in the captions generated by the \modelName{} model.}
    \label{fig:imgnet_noc_errors_short}
\end{figure}

\begin{figure*}[!htb]
\vspace{-0.3cm}
\centering
\includegraphics[scale=0.44]{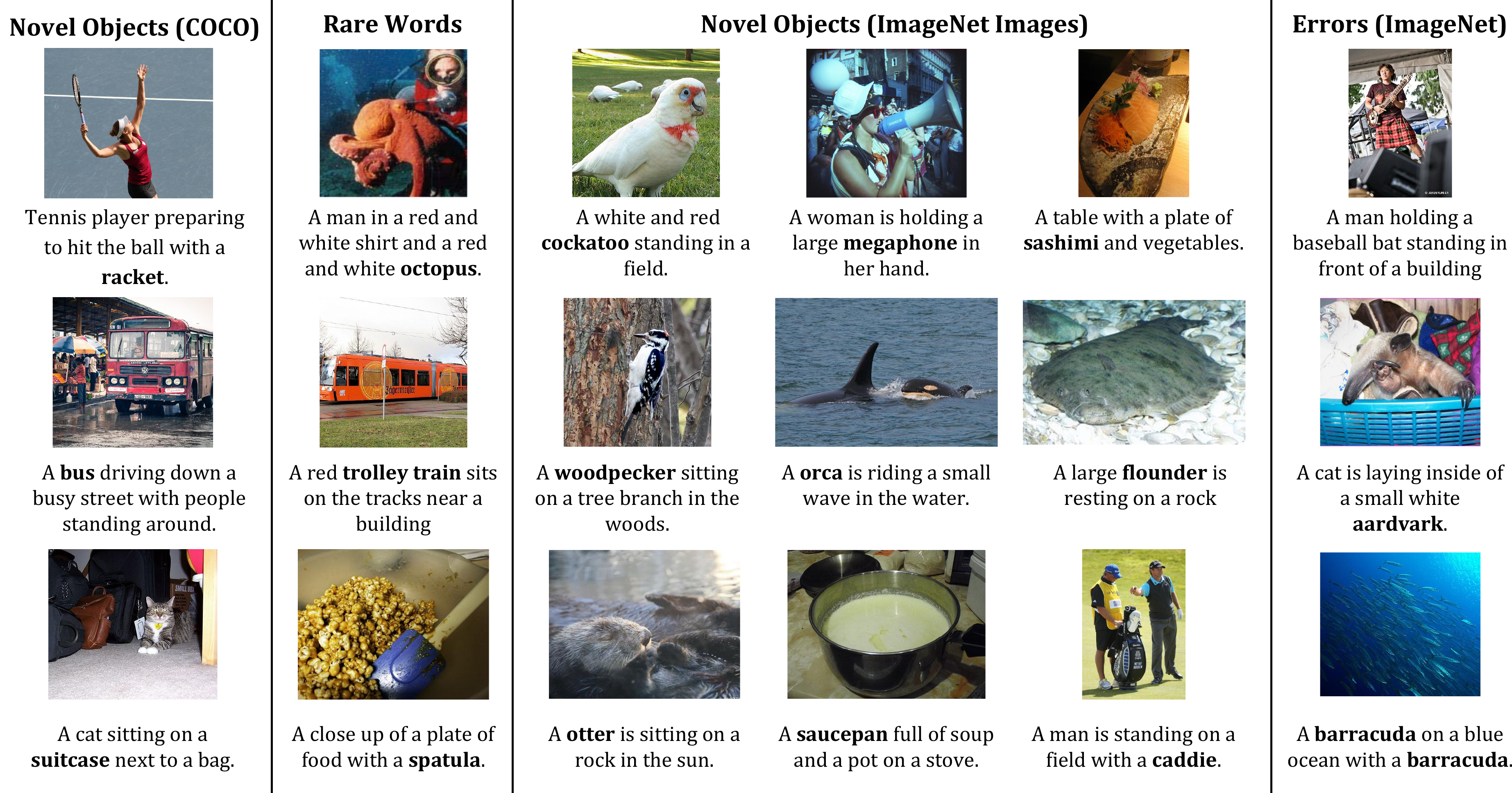}
 \caption{\small
Descriptions produced by \modelName{} on a variety of objects, including ``caddie'', ``saucepan'', and ``flounder''.
(Right) \modelName{} makes errors and (top right) fails to describe the new object (``sitar''). More categories of images and objects are in the supplement.}
\label{fig:imagenet_examples}
\end{figure*}

Fig.~\ref{fig:imnet_dcc_noc} and Fig.~\ref{fig:imagenet_examples} (column 3) show examples where
\modelName{} describes a large variety of objects from ImageNet. Fig.~\ref{fig:imnet_dcc_noc} compares our model with DCC.
Fig.~\ref{fig:imgnet_noc_errors_short} and Fig.~\ref{fig:imagenet_examples} (right) outline some errors.
Failing to describe a new object is one common error for \modelName{}.
E.g. Fig.~\ref{fig:imagenet_examples} (top right), \modelName{} incorrectly describes a man holding a ``sitar'' as a man holding a ``baseball bat''.
Other common errors include generating non-grammatical or nonsensical phrases (example with ``gladiator'', ``aardvark'') %
 and repeating a specific object (``A barracuda ... with a barracuda'', ``trifle cake''). 

\vspace{-0.1cm}
\subsection{Describing Rare Objects/Words}
\vspace{-0.1cm}
The selected rare words occur with varying frequency in the MSCOCO training set, with about 52 mentions on average (median 27) across all training sentences. For example, words such as ``bonsai'' only appear 5 times,``whisk'' (11 annotations), ``teapot'' (30 annotations), and others such as pumpkin appears 58 times, ``swan'' (60 annotations), and on the higher side objects like scarf appear 144 times.
When tested on ImageNet images containing these concepts, a model trained only with MSCOCO paired data incorporates rare words into sentences 2.93\% of the time with an average F1 score of 4.58\%.
In contrast, integrating outside data, our \modelName{} model can incorporate rare words into descriptions 35.15\% of the time with an average F1 score of 47.58\%. We do not compare this to DCC since DCC cannot be applied directly to caption rare objects.

\subsection{Human Evaluation}
ImageNet images do not have accompanying captions and this makes the task much more challenging to evaluate. To compare the performance of \modelName{} and DCC we obtain human judgements on captions generated by the models on several object categories. We select 3 images each from about 580 object categories that at least one of the two models, \modelName{} and DCC, can describe. (Note that although both models were trained on the same ImageNet object categories, \modelName{} is able to describe almost all of the object categories that have been described by DCC). When selecting the images, for object categories that both models can describe, we make sure to select at least two images for which both models mention the object label in the description. Each image is presented to three workers. We conducted two human studies (sample interface is in the supplement): Given the image, the ground-truth object category (and meaning), and the captions generated by the models, we evaluate on:
\vspace{-0.2cm}
\begin{description}
\setlength{\topsep}{0pt}
\setlength\itemsep{-0.4em}
\item[Word Incorporation:] We ask humans to choose which sentence/caption incorporates the object label meaningfully in the description. The options provided are: (i) Sentence 1 incorporates the word better, (ii) Sentence 2 incorporates the word better, (iii) Both sentences incorporate the word equally well, or (iv) Neither of them do well.
\item[Image Description:] We also ask humans to pick which of the two sentences describes the image better.
\end{description}
\vspace{-0.1cm}
This allows us to compare both how well a model incorporates the novel object label in the sentence, as well as how appropriate the description is to the image. The results are presented in Table \ref{tab:imgnet-human}. On the subset of images corresponding to objects that both models can describe (Intersection), \modelName{} and DCC appear evenly matched, with \modelName{} only having a slight edge. However, looking at all object categories (Union), \modelName{} is able to both incorporate the object label in the sentence, and describe the image better than DCC.

%% file: 7conclusion.tex
\vspace{-0.1cm}
\section{Conclusion}
\vspace{-0.2cm}
We present an end-to-end trainable architecture that incorporates auxiliary training objectives and distributional semantics to generate descriptions for object classes unseen in paired image-caption data. Notably, \modelName{}'s architecture and training strategy enables the visual recognition network to retain its ability to recognize several hundred categories of objects even as it learns to generate captions on a different set of images and objects.
We demonstrate our model's captioning capabilities on a held-out set of MSCOCO objects as well as several hundred ImageNet objects. Both human evaluations and quantitative assessments show that our model is able to describe many more novel objects compared to previous work. \modelName{} has a 10\% higher F1 on unseen COCO objects and 20\% higher F1 on ImageNet objects compared to previous work, while also maintaining or improving descriptive quality. We also present an analysis of the contributions from different network modules, training objectives, and data sources. Additionally, our model directly extends to generate captions for ImageNet objects mentioned rarely in the image-caption corpora. 
Code is available at: \href{https://vsubhashini.github.io/noc.html}{\footnotesize\tt{https://vsubhashini.github.io/noc.html}}
\begin{table}
\small
\begin{center}
\begin{tabular}{lcccc}
\toprule
& \multicolumn{2}{c}{{Word Incorporation}} & \multicolumn{2}{c}{{Image Description}} \\
Objects subset $\rightarrow$ & {\scriptsize{Union}} & {\scriptsize{Intersection}} & {\scriptsize{Union}} & {\scriptsize{Intersection}} \\
\cmidrule(lr){1-5}
NOC is better & \textbf{43.78} & 34.61 & \textbf{59.84} & 51.04 \\
DCC is better & 25.74 & 34.12 & 40.16 & 48.96 \\
Both equally good & \ \ 6.10 & \ \ 9.35 &  \multicolumn{2}{c}{-} \\
Neither is good &  24.37 & 21.91 &  \multicolumn{2}{c}{-} \\
\bottomrule
\end{tabular}
\end{center}
\caption{\small ImageNet: Human judgements comparing our \modelName{} model with DCC \protect\cite{hendricks16cvpr} on the ability to meaningfully incorporate the novel object in the description (Word Incorporation) and describe the image. `Union' and `Intersection' refer to the subset of objects where atleast one model, and both models are able to incorporate the object name in the description. All values in $\%$.}
\label{tab:imgnet-human}
\end{table}

%% file: 8supplement.tex
\appendix
\section*{Supplement}

This supplement presents further qualitative results of our Novel Object Captioner (\modelName{}) model on Imagenet images (in Sec.~\ref{sec:imnet-qual}), details pertaining to the quantitative results on COCO held-out objects (in Sec.~\ref{sec:coco-quant}), as well as the interface used by Mechanical Turk workers comparing \modelName{} with prior work (in Sec.~\ref{sec:turk-interface}).

\section{ImageNet Qualitative Examples}
\label{sec:imnet-qual}
We present additional examples of the \modelName{} model's descriptions on Imagenet images. We first present some examples where the model is able to generate descriptions of an object in different contexts. Then we present several examples to demonstrate the diversity of objects that \modelName{} can describe. We then present examples where the model generates erroneous descriptions and categorize these errors.

\subsection{Context}
\begin{figure*}[htb]
\vspace{2.2cm}
\centering
\includegraphics[scale=0.65]{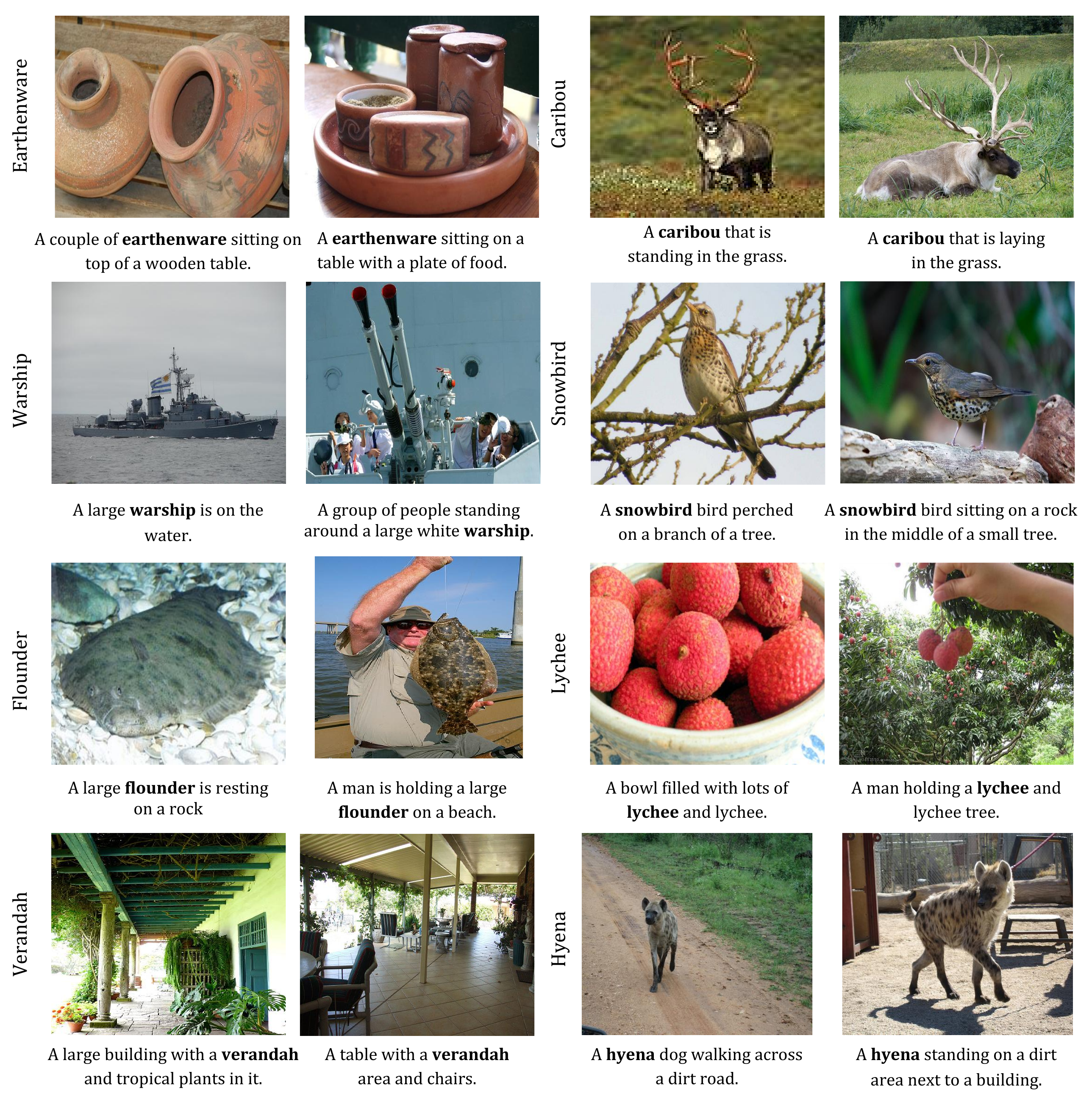}
\caption{Examples showing descriptions generated by \modelName{} for ImageNet images of eight objects, each in two different contexts. \modelName{} is often able to generate descriptions incorporating both the novel object name as well as the background context correctly.}
 \label{fig:supp_context}
\vspace{2.2cm}
\end{figure*}

Fig.~\ref{fig:supp_context} shows images of eight objects, each in two different settings from ImageNet. Images show objects in different backgrounds (Snowbird on a tree branch and on a rock, Hyena on a dirt path and near a building); actions (Caribou sitting vs lying down); and being acted upon differently (Flounder resting and a person holding the fish, and Lychees in a bowl vs being held by a person). \modelName{} is able to capture the context information correctly while describing the novel objects (eartherware, caribou, warship, snowbird, flounder, lychee, verandah, and hyena).

\subsection{Object Diversity}
\begin{figure*}[htb]
\vspace{-0.5cm}
\centering
\includegraphics[scale=0.62]{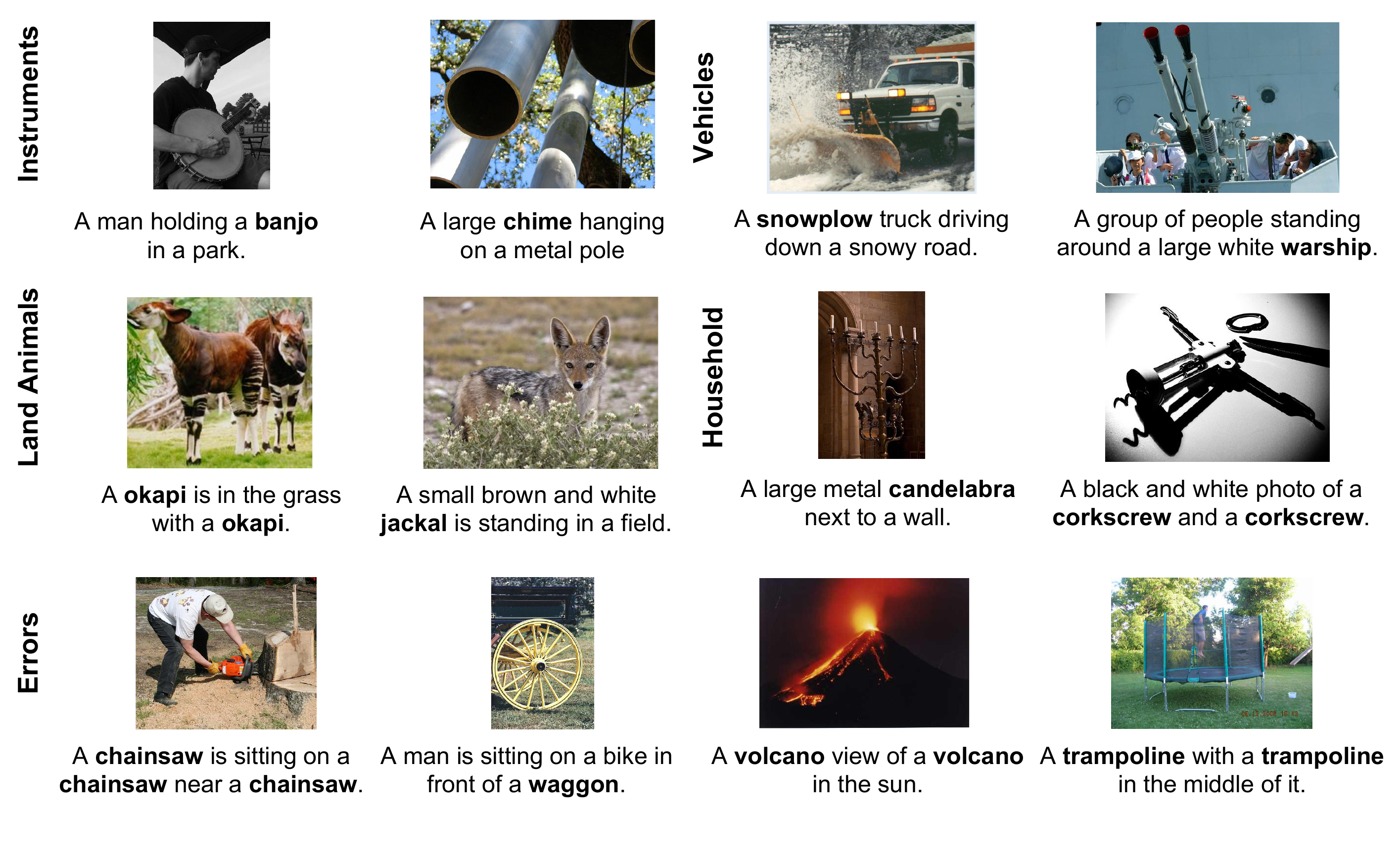}
\vspace{-1.0cm}
\caption{Examples of sentences generated by our \modelName{} model on ImageNet images of objects belonging to a diverse variety of categories including food, instruments, outdoor scenes, household equipment, and vehicles. The novel objects are in \textbf{bold}.
The last row highlights common errors where the model tends to repeat itself or hallucinate objects not present in the image.}
\label{fig:supp_noc_1}
\end{figure*}

\begin{figure*}[htb]
\vspace{-0.3cm}
\centering
\includegraphics[scale=0.57]{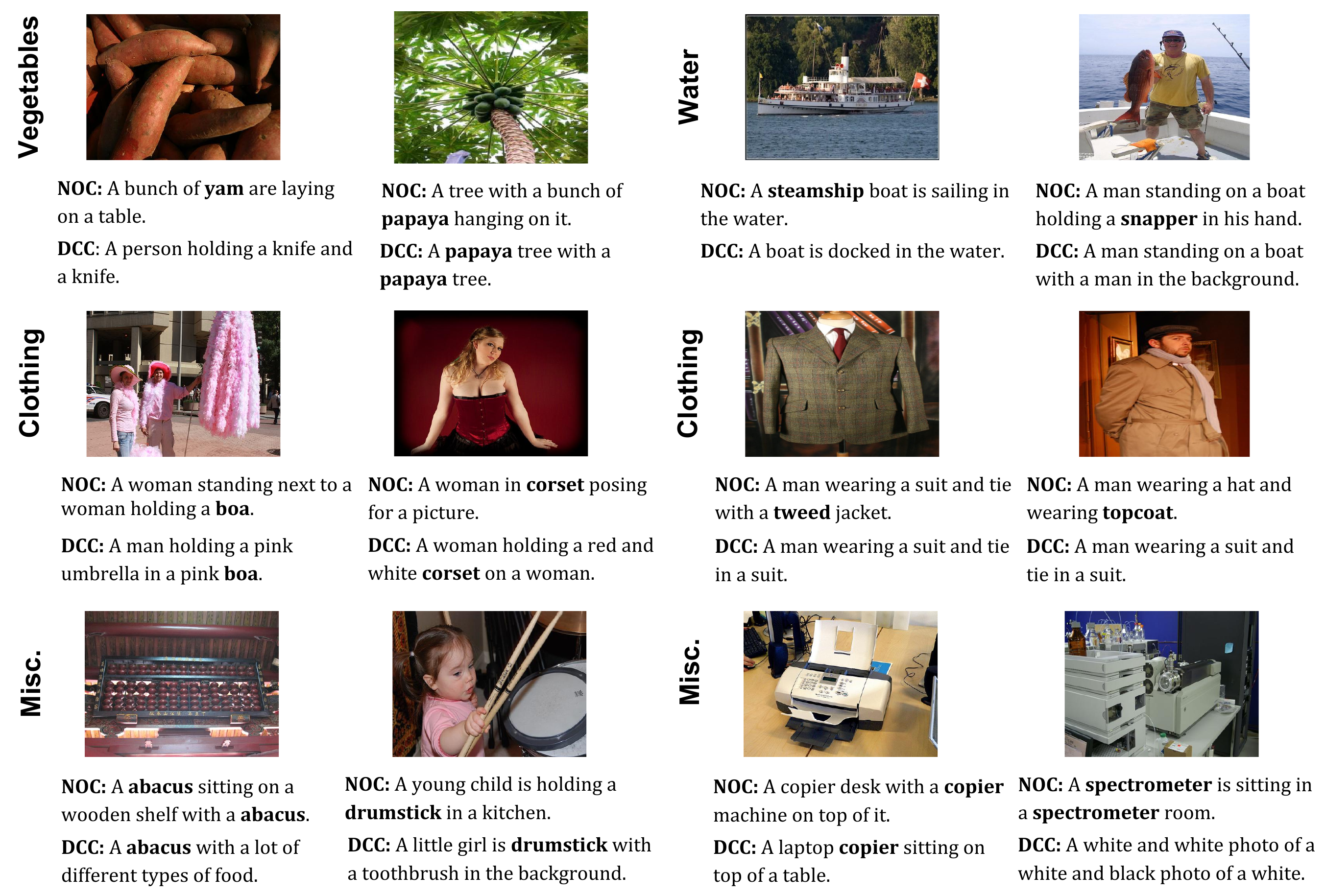}
\caption{Examples comparing sentences generated by DCC \cite{hendricks16cvpr} and our \modelName{} model on ImageNet images of object categories that \textit{both} models can describe including food, animals, vegetables/fruits, indoor and outdoor scenes, and clothing. The novel objects are in \textbf{bold}.}
\label{fig:supp_noc_2}
\end{figure*}

Fig.~\ref{fig:supp_noc_1} and Fig.~\ref{fig:supp_noc_2} present descriptions generated by \modelName{} on a variety of object categories such as birds, animals, vegetable/fruits, food items, household objects, kitchen utensils, items of clothing, musical instruments, indoor and outdoor scenes among others. While almost all novel words (nouns in Imagenet) correspond to objects, \modelName{} learns to use some of them more appropriately as adjectives (`chiffon' dress in Fig.~\ref{fig:supp_noc_1}, `brownstone' building and `tweed' jacket in Fig.~\ref{fig:supp_noc_2} as well as `woollen' yarn in 
Fig.~\ref{fig:imnet_dcc_noc}.
\paragraph{Comparison with prior work.} Additionally, for comparison with the DCC model from \cite{hendricks16cvpr}, Fig.~\ref{fig:supp_noc_2} presents images of objects that both models can describe, and captions generated by both DCC and \modelName{}.

\begin{figure*}[htb]
\centering
\includegraphics[scale=0.62]{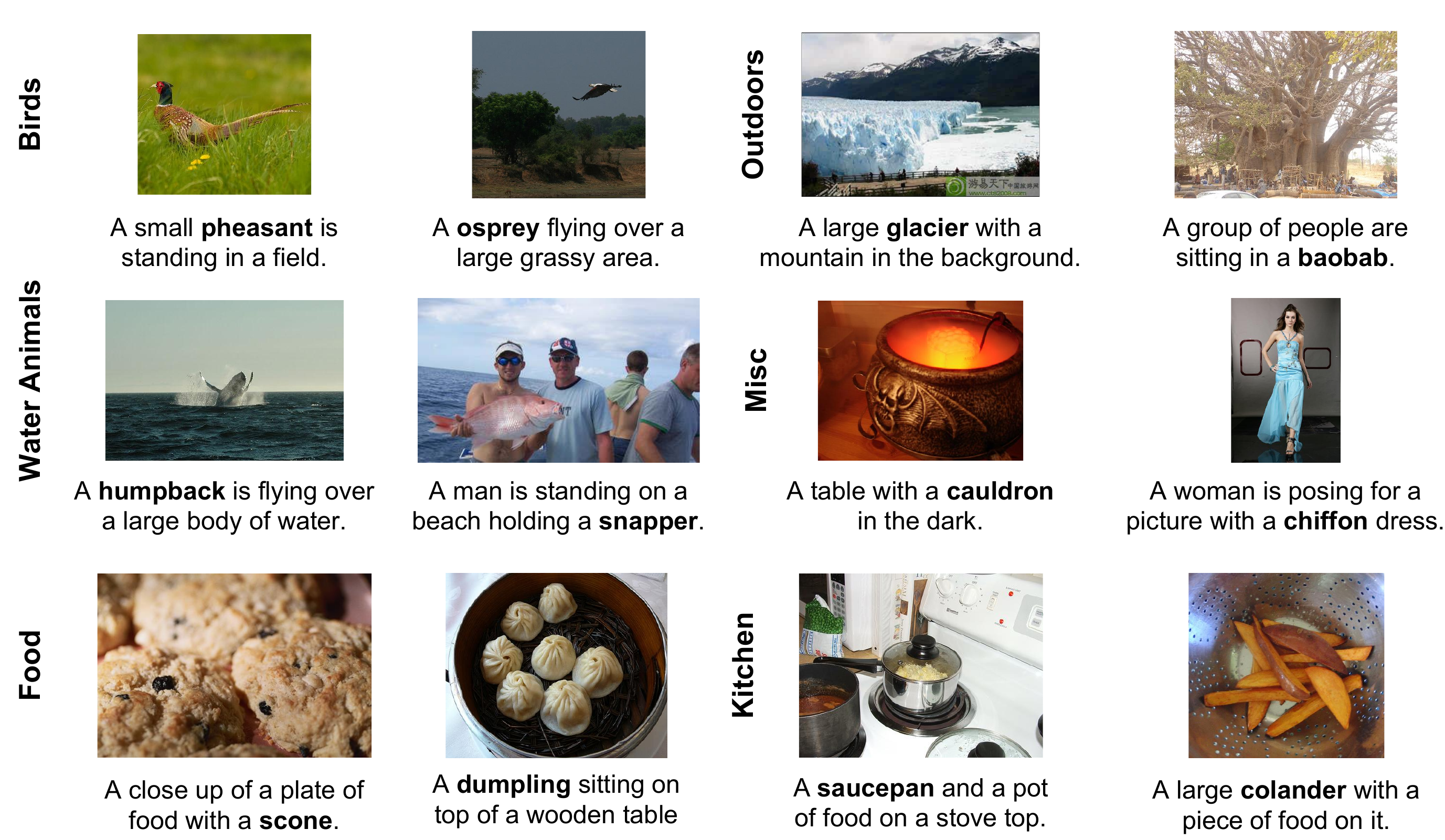}
\vspace{-0.4cm}
\end{figure*}

\begin{figure*}[htb]
\centering
\vspace{-0.4cm}
\includegraphics[scale=0.57]{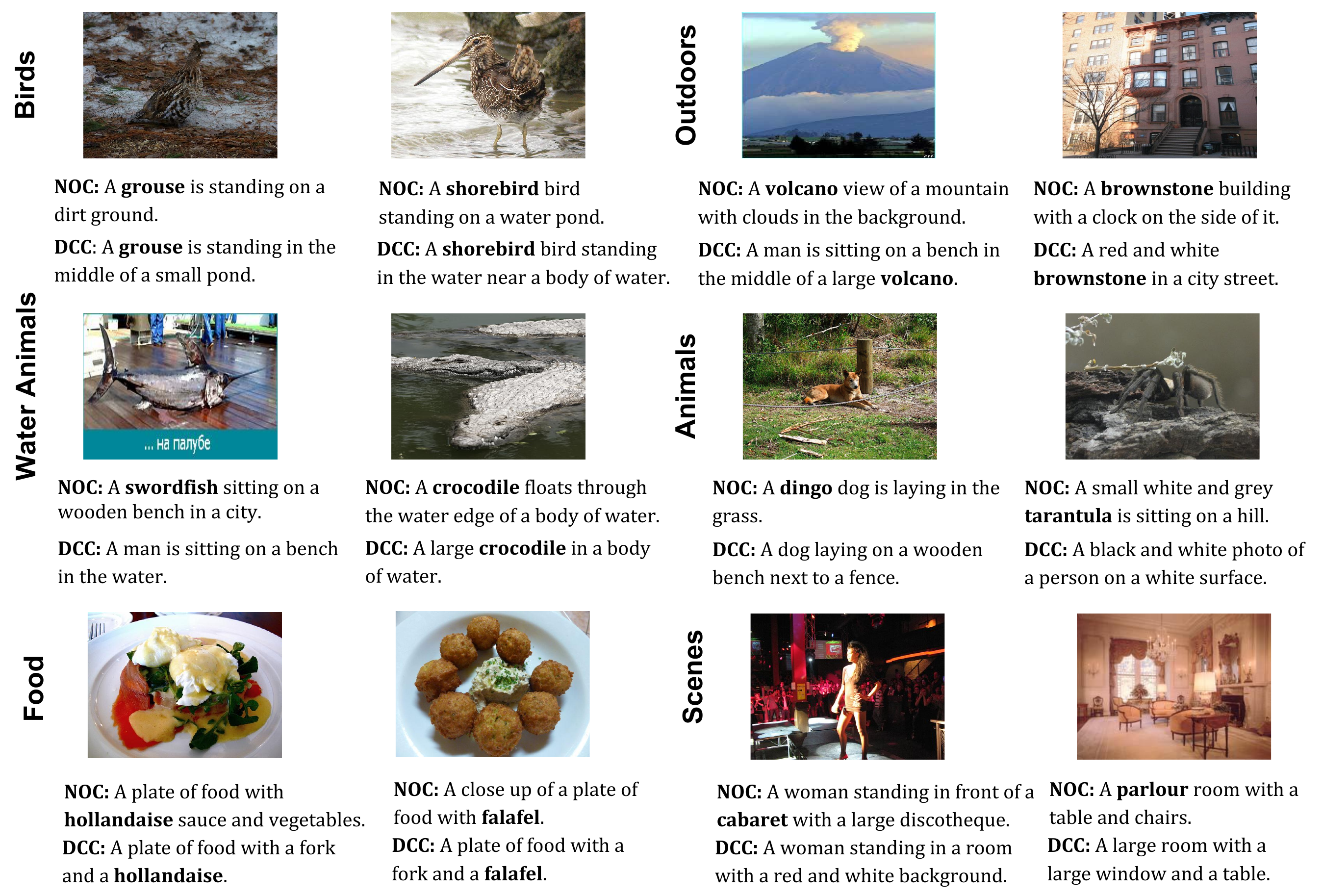}
\vspace{-0.4cm}
\end{figure*}

\subsection{Categorizing Errors}
\begin{figure*}[htb]
\centering
\includegraphics[scale=0.71]{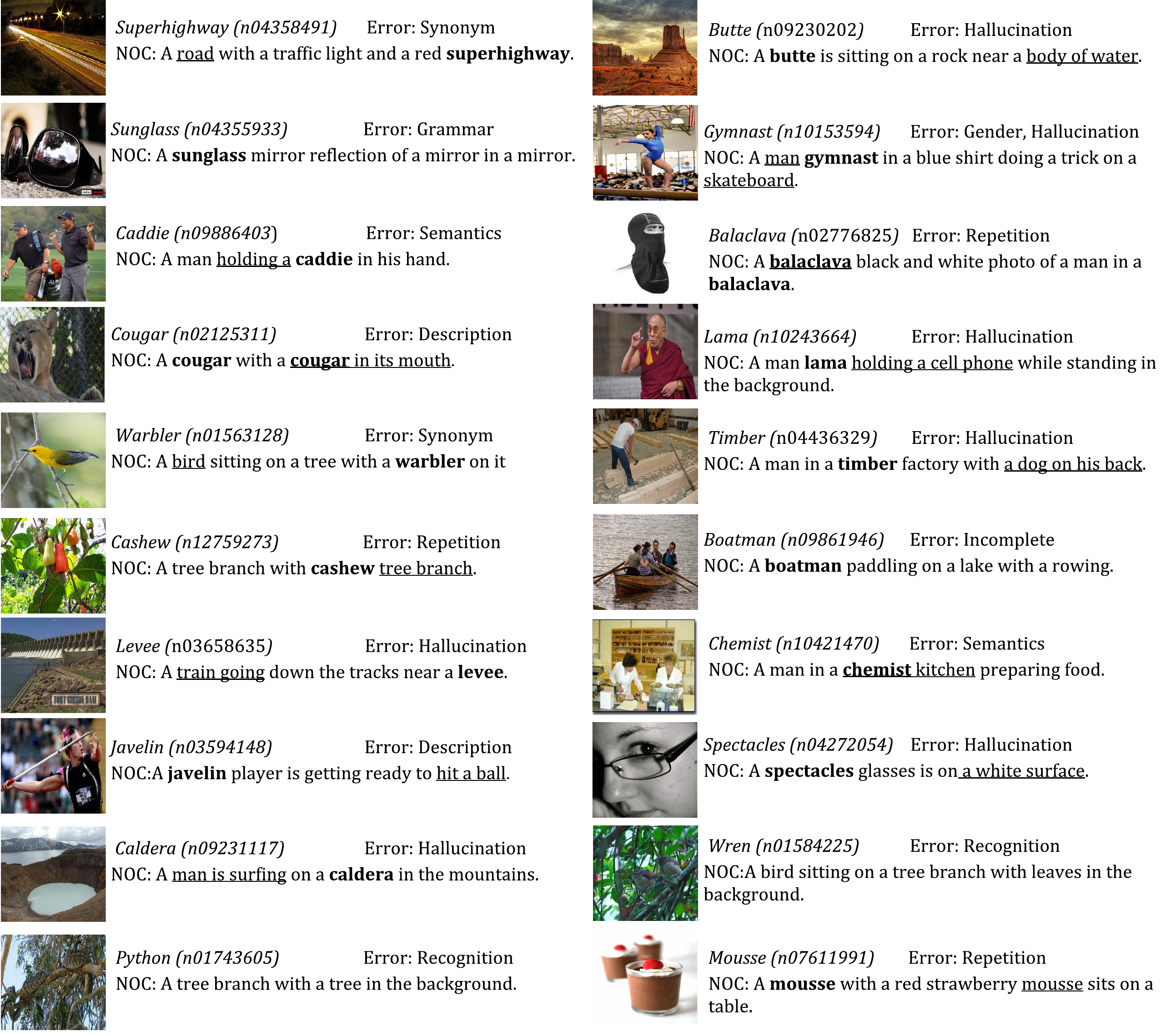}
\vspace{0.3cm}
\caption{Examples of images where the model makes errors when generating descriptions. The novel object is in \textbf{bold} and the {\uline{errors are underlined}}. \modelName{} often tends to repeat words in its description, or hallucinate objects not present in the image. The model sometime misidentifies gender, misrepresents the semantics of the novel object, or just makes grammatical errors when composing the sentence.}
 \label{fig:supp_lemons}
\end{figure*}

Fig.~\ref{fig:supp_lemons} presents some of the errors that our model makes when captioning Imagenet images. While \modelName{} improves upon existing methods to describe a variety of object categories, it still makes a lot of errors. The most common error is when it simply fails to recognize the object in the image (e.g. image with `python') or describes it with a more generic hyponym word (e.g. describing a bird species such as `wren' or `warbler' in Fig.~\ref{fig:supp_lemons} as just `bird').
For objects that the model is able to recognize, the most common errors are when the model tends to repeat words or phrases (e.g. descriptions of images with `balaclava', `mousse' and `cashew'), or just hallucinate other objects in the context that may not be present in the image (e.g. images with `butte', `caldera', `lama', `timber'). 
Sometimes, the model does get confused between images of other similar looking objects (e.g. it confuses `levee' with `train'). Apart from these the model does make mistakes when identifying gender of people (e.g. `gymnast'), or just fails to create a coherent correct description even when it identifies the object and the context (e.g. images of `sunglass' and `cougar').

\begin{figure*}[htb]
\centering
\includegraphics[scale=0.65]{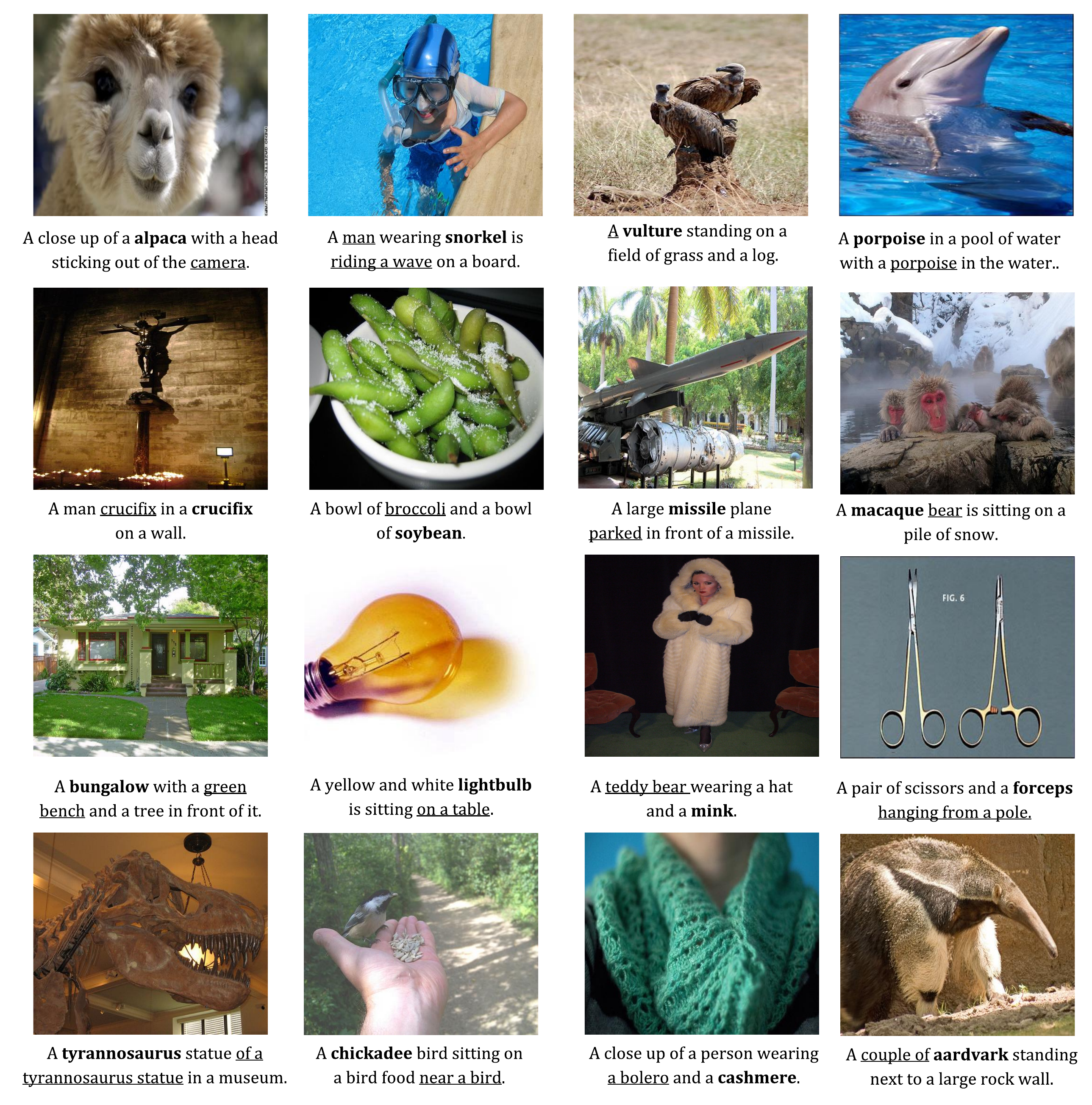}
\caption{Some examples where \modelName{} makes minor errors when describing the image. The novel object is in \textbf{bold} and the {\uline{word or segment corresponding to the error is underlined}.} Counting, repetitions, confusing object categories (e.g. `macaque', `bear'), grammatical errors, and hallucinating objects that are absent are some common errors that the model makes. However, the generated description is still meaningful and relevant.}
 \label{fig:supp_minor_errors}
\end{figure*}

\paragraph{Relevant but Minor Errors.} 
Fig.~\ref{fig:supp_minor_errors} presents more examples where \modelName{} generates very relevant descriptions but makes some minor errors with respect to counting (e.g. images of `vulture' and `aardvark'), age (e.g. refers to boy wearing `snorkel' as `man'), confusing the main object category (e.g. `macaque' with `bear' and person as `teddy bear') or makes minor word repetitions, and grammatical errors.

\section{MSCOCO Quantitative Results}
\label{sec:coco-quant}
We present detailed quantitative results comparing DCC and NOC on the 8 held-out objects.

\subsection{F1 and METEOR}
While Table.\ \ref{tab:results:coco-8objs} %
presents the F1 scores comparing the DCC model \cite{hendricks16cvpr} and our \modelName{} model for each of the eight held-out objects in the test split,  Table.~\ref{tab:results:coco-8objs-meteor} supplements  this by also providing the individual meteor scores for the sentences generated by the two models on these eight objects. In case of NOC, we sampled sentences (25) and picked one with lowest log probability. Using, beam search with a beam width of 1 produces sentences with METEOR score 20.69 and F1 of 50.51. In Tables ~\ref{tab:noc-comp} and ~\ref{tab:ablations}, all lines except the last line corresponding to NOC use beam search with a beam-width of 1.

\begin{table*}[htb]
\small
\centering
\begin{center}
\begin{tabular}{clccccccccc}
\toprule
Metric & Model &  bottle & bus & couch & microwave & pizza & racket & suitcase & zebra & Avg. \\
\cmidrule(lr){1-11}
\multirow{2}{*}{F1} & DCC & 4.63 & 29.79 & \textbf{45.87} & \textbf{28.09} & 64.59 & 52.24 & 13.16 & 79.88 & 39.78 \\
& \modelName{} (ours) & \textbf{17.78} & \textbf{68.79} & 25.55 & 24.72 & \textbf{69.33} & \textbf{55.31} & \textbf{39.86 }& \textbf{89.02} & {\bf{48.79}} \\
\cmidrule(lr){1-11}
\multirow{2}{*}{METEOR} & DCC & 18.1 & \textbf{21.6}& \textbf{23.1} & 22.1 & 22.2 & 20.3 & 18.3 & 22.3 & 21.00 \\
& \modelName{} (ours) & \textbf{21.2} & 20.4 & 21.4 & 21.5 & 21.8 & \textbf{24.6} & 18.0 & 21.8 & \textbf{21.32}\\
\bottomrule
\end{tabular}
\vspace{0.3cm}
\caption{
MSCOCO Captioning: F1 and METEOR scores (in \%) of \modelName{} (our model) and DCC \protect\cite{hendricks16cvpr} on the held-out objects not seen jointly during image-caption training, along with the average scores of the generated captions across images containing these objects.
}
\label{tab:results:coco-8objs-meteor}
\end{center}
\end{table*}

\subsection{Word-embedding for DCC and NOC}
One aspect of difference between NOC and DCC is that NOC uses GloVe embeddings in it's language model whereas DCC uses word2vec embeddings to select similar objects for transfer. In order to make a fair comparison of DCC with NOC, it is also important to consider the setting where both models use the same word-embedding. We modify the transfer approach in DCC and replace word2vec with GloVe embeddings. From Table.~\ref{tab:glove-stats} we note that the difference in DCC is not significant. Thus, the embeddings themselves do not play as significant a role as the joint training approach.
\begin{table}[htb]
\begin{center}
\small
\begin{tabular}{| l | c | c |}
  \hline
   Model & F1 (\%) & METEOR (\%) \\ \hline
  DCC with word2vec & 39.78 & 21.00 \\
  DCC with GloVe & 38.04 & 20.26  \\
  NOC (ours, uses GloVe) & \textbf{48.79} & \textbf{21.32} \\
  \hline
\end{tabular}
\end{center}
\vspace{-0.2cm}
\caption{DCC and NOC both using GloVe on MSCOCO dataset.}
\label{tab:glove-stats}
\end{table}

\subsection{Joint Training with Auxiliary Objectives}
When performing joint training and considering the overall optimization objective as the sum of the image-specific loss, the text-specific loss and image-caption loss, we can define the objective more generally as:
\vspace{-0.2cm}
\begin{align}
\vspace{-0.2cm}
\label{eqn:jointLgen}
\mathcal{L} &= \mathcal{L_{CM}} + \alpha \mathcal{L_{IM}} + \beta \mathcal{L_{LM}}
\end{align}
where $\alpha$ and $\beta$ are hyper-parameters which determine the weighting between different losses. In our experiments setting $\alpha=1$ and $\beta=1$ provided the best performance on the validation set. Other values of $(\alpha, \beta) \in \{(1, 2), (2, 1)\}$ resulted in lower F1 and METEOR scores.

\section{Mechanical Turk Interface}
\label{sec:turk-interface}
\begin{figure*}[!htbp]
\centering
\begin{center}
\includegraphics[scale=0.44]{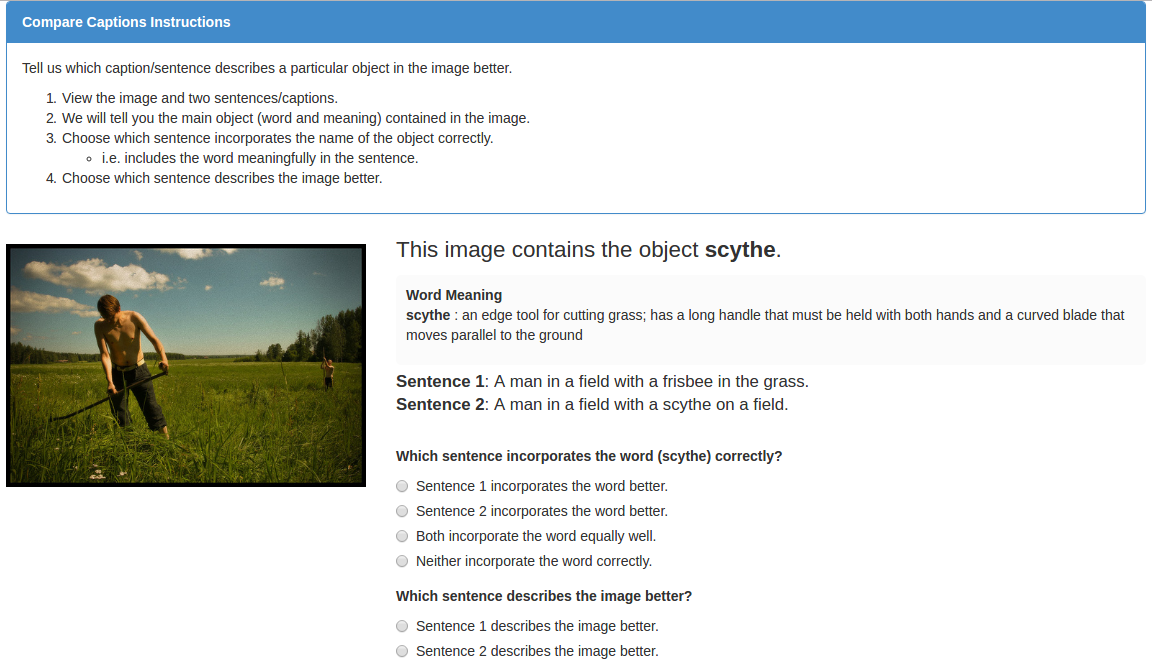}
\caption{Interface used by mechanical turk workers when comparing captions/sentences generated by our \modelName{} model with previous work (DCC \cite{hendricks16cvpr}). The workers are asked to compare on both Word Incorporation i.e. how well each model incorporates the novel object in the sentence, as well as Image Description i.e. which caption describes the image better.}
 \label{fig:supp_turk_ui}
\end{center}
\end{figure*}

Fig.~\ref{fig:supp_turk_ui} presents the interface used by mechanical turk workers when comparing sentences generated by our model and previous work. The workers are provided with the image, the novel object category (word as well as meaning) that is present in the image, and two sentences (one each from our model and previous work). The sentence generated by the \modelName{} model is randomly chosen to be either Sentence 1 or Sentence 2 for each image (with the other sentence corresponding to the caption generated by previous work \cite{hendricks16cvpr}). Three workers look at each image and the corresponding descriptions. The workers are asked to judge the captions based on how well it incorporates the novel object category in the description, and which sentence describes the image better.

\section{Future directions}
One interesting future direction would be to create a model that can learn on new image-caption data after it has already been trained. This would be akin to \cite{mao15iccv}, where after an initial NOC model has already been trained we might want to add more objects to the vocabulary, and train it on few image-caption pairs. The key novelty would be to improve the captioning model by re-training only on the new data instead of training on all the data from scratch.